\pdfoutput=1

\documentclass[11pt]{article}

\usepackage{float}
\usepackage{acl}

\usepackage{times}
\usepackage{latexsym}
\usepackage{amsmath}
\usepackage{amssymb}
\usepackage[T1]{fontenc}
\usepackage{graphicx}

\usepackage[utf8]{inputenc}

\usepackage{microtype}

\title{Evaluating the Deductive Competence of Large Language Models}

\author{Spencer M. Seals\textsuperscript{1,2,3} \and Valerie L. Shalin\textsuperscript{1,4} \\
        \textsuperscript{1}Wright State University \textsuperscript{2}Oak Ridge Institute for Science and Education \\\textsuperscript{3}Air Force Research Laboratory \textsuperscript{4}AI Institute - University of South Carolina \\
        \texttt{s.m.seals@outlook.com}}

\begin{document}
\maketitle

\begin{abstract}
The development of highly fluent large language models (LLMs) has prompted increased interest in assessing their reasoning and problem-solving capabilities. We investigate whether several LLMs can solve a classic type of deductive reasoning problem from the cognitive science literature. The tested LLMs have limited abilities to solve these problems in their conventional form. We performed follow up experiments to investigate if changes to the presentation format and content improve model performance. We do find performance differences between conditions; however, they do not improve overall performance. Moreover, we find that performance interacts with presentation format and content in unexpected ways that differ from human performance. Overall, our results suggest that LLMs have unique reasoning biases that are only partially predicted from human reasoning performance and the human-generated language corpora that informs them.
\end{abstract}

\section{Introduction}
\label{sec:introduction}

\footnote{https://github.com/spencer-michael-s/deductive-competence} \footnote{The views expressed are those of the author and do not necessarily reflect the official policy or position of the Department of the Air Force, the Department of Defense, or the U.S. government.} The development and availability of highly fluent large language models (LLMs) (i.e., \citep{brown_language_2020, devlin-etal-2019-bert, ouyangTrainingLanguageModels2022, zhangOPTOpenPretrained2022}) has increased interest in assessing their reasoning and problem solving abilities \citep{bhargava_commonsense_2022, geva-etal-2020-injecting, jumelet-etal-2019-analysing, mitchell_abstraction_2021, trinh_simple_2019, webb_emergent_2022}. Despite considerable performance improvements on benchmark tasks, LLMs exhibit mixed results on reasoning tasks. Some research has suggested that LLMs may have emergent reasoning abilities that enable better performance than those of human subjects \citep{webb_emergent_2022}. Other research has suggested that LLM reasoning performance is inconsistent and task dependent. Such research has suggested that some tasks, such as four term analogy problems \citep{mikolov-etal-2013-linguistic} and different natural language inference tasks \citep{ williams_broad-coverage_2018}, are simply easier to solve. Other types of reasoning tasks such as analogy generation \citep{bhavya_analogy_2022} and deductive competence \citep{dasgupta_language_2022} are more challenging.

\citep{dasgupta_language_2022} has investigated deductive competence in LLMs with characteristically mixed results. They demonstrated that one LLM, Chinchilla \citep{hoffmann_training_2022}, showed content effects on reasoning similar to human behavior documented in the cognitive science literature. For zero-shot performance, they found 50\% accuracy for what they call realistic problems but chance accuracy for unrealistic problems. A 5-shot presentation resulted in some performance improvement for realistic problems, but performance on unrealistic problems remained low.

In this paper, we extend the previous research in several ways. First, we investigate the extent to which limited performance may be due to how the task was formatted. Prior research has demonstrated that overall performance can vary according to how a particular task is formatted \citep{gao-etal-2021-making,jiang-etal-2021-know,li-liang-2021-prefix,shin-etal-2020-autoprompt}. Research on analogy generation \citep{bhavya_analogy_2022} demonstrates that performance depends on the specific prompt given to the models. Thus the inability of a model to solve one particular format of a task only provides a lower limit for assessing whether a model can successfully solve that task \citep{jiang-etal-2021-know}.

Second, limited performance on deductive reasoning may be due to how the researchers employed content familiarity, of direct relevance to the distribution of content in the training corpus. For familiar content, they tested several different types of problems including social rules (i.e. \textit{If people are driving, they must have a license}) and other relationships (i.e., \textit{If the plants are flowers, they must be fertilized}). However, content familiarity does not fully capture the content benefit seen in human subjects \citep{griggs_elusive_1982, manktelow_facilitation_1979}. Instead, previous research indicates that people perform substantially better on problems that involve social rules than those that do not \citep{cosmides_logic_1989}, even when the problems contain other types of familiar content \citep{griggs_elusive_1982}.

We expand the research base on the reasoning capabilities of LLMs by: 1) examining the role of specifically social-rules in reasoning about realistic content, 2) investigating the role of alternative presentation formats in deductive reasoning performance, and 3) expanding the set of candidate LLMs to evaluate potential algorithmic effects. Our results show that social content does benefit LLM performance, but not to the extent that might be expected based on a human sourced training corpus.  While presentation formats do influence performance, they interact with content in a surprising (non-human) fashion. These findings are independent of the LLM examined.

\section{Evaluating Deductive Competence}
The Wason selection task is a reasoning task from the cognitive science literature that evaluates deductive competence \citep{wason_reasoning_1968}. Participants are presented with a rule of the form \textit{If p, then q} and four cards with p status on one side and q status on the other that correspond to the options  $P, \neg P, Q,$ and $\neg Q$. Participants are asked to determine which card or cards must be flipped over to validate whether the rule holds for this set of cards.

\begin{table}
\centering
\begin{tabular}{ll}
\hline
\textbf{Inference} & \textbf{Definition}\\
\hline
    \textit{Modus ponens} & $P \implies Q$, $P \therefore Q$ \\
    Deny the antecedent & $P \implies Q$, $\neg P \therefore \neg Q$ \\
    Affirm the consequent & $P \implies Q, Q \therefore P$ \\
    \textit{Modus tollens} & $P \implies Q, \neg Q \therefore \neg P$  \\\hline
\end{tabular}
\caption{Four conditional inferences in the Wason Task}
\label{tab:cond_inference}

\end{table}

In the traditional \textit{abstract} version of the task, participants are given rules about letters and numbers (i.e., \textit{if there is an odd number on one side of the card, there is a vowel on the other side of the card}). The correct response requires the identification of \textit{two} cards. Typical human accuracy for these problems is 10-20\% with common errors consistent with confirmation bias. In contrast, problems that deal with a social rule (i.e., \textit{if a person is drinking beer, they must be at least 21 years old}) are easy to solve- most participants (70\%+) correctly select both cards \citep{griggs_elusive_1982}.

There are four potential conditional inferences in task: \textit{modus ponens}, denial of the antecedent, affirmation of the consequent, and \textit{modus tollens} \citep{Evans2013}. Of these four inferences, only \textit{modus ponens} and \textit{modus tollens} are logically valid. These inferences and their logical forms are illustrated in Table~\ref{tab:cond_inference}. 

The Wason task makes a good candidate task for evaluating the reasoning performance of LLMs for several reasons. First, the task is relatively close to certain language modeling objectives. While the task involves a reasoning component, it can be formatted as a completion task, where the objective is to predict the answer given the problem text. This suggests that prior training, particularly for LLMs with high numbers of parameters that have been trained on large text corpora, should provide sufficient information for performing the Wason task. 
Moreover, the construction of the task minimizes the potential for confounds that may artificially inflate performance \citep{hovy-yang-2021-importance, mitchell_debate_2023, rudinger-etal-2017-social}. Previous work has demonstrated that high performance on some natural language inference tasks \citep{bowman-etal-2015-large, williams_broad-coverage_2018} can be due to exploitable properties of the training data  \citep{gururangan-etal-2018-annotation}. The standardized format of the Wason task allows for the creation of a large number of carefully constructed examples without the risk that some answers may be easily determined from the original prompt alone.

Second, the problem examines both straightforward and challenging aspects of deductive competence. As noted above, a correct answer to a Wason problem involves two logical processes: \textit{modus ponens} and \textit{modus tollens}. Results from the cognitive science literature indicate that these rules are not equally difficult- applying \textit{modus ponens} is considerably easier than applying \textit{modus tollens}. The vast majority of people correctly apply \textit{modus ponens}, even for difficult problems (i.e., \citep{wason_reasoning_1968, griggs_elusive_1982, manktelow_facilitation_1979}. In contrast, people fail to apply \textit{modus tollens}, unless the problem has a particular form of semantic content associated with it. Similarly, we might expect presentation format to assist LLMs. 

Third, the way the task is constructed allows for careful examination of how problem content influences reasoning performance. Because LLMs are built on co-occurring content in human-sourced corpora, they should benefit from problems that contain familiar relationships. This should be especially true for problems where the relationship between the antecedent and the consequent is highly familiar. For example, in the rule \textit{If a person is driving, they must have a driver's license}, the antecedent \textit{driving a car}, and the consequent \textit{driver's license} have a familiar (and commonly occurring) relationship. In comparison, the antecedent and the consequent in a rule such as \textit{If a person is driving, they must have a book bag} do not have the same familiar relationship. While it is likely that some problems may be more difficult than others (i.e., because some completions are more probable) we control for this experimentally. We create sets of problems where both arguments contain realistic content, but the relationship between them is unfamiliar (see Appendix ~\ref{app:ex_problems}). 

Lastly, there is a large body of human performance literature on this task. This literature provides a comparison point for evaluating the performance of LLMs.

\section{Experiments}
In this section, we discuss the conditions, task format, models, and evaluation metrics associated with our experiments.

\subsection{Task Conditions}

We evaluate a total of 350 problems, 325 of which we created for this project. The remaining 25 were drawn from recent work \cite{dasgupta_language_2022} and sorted into our problem categories. To facilitate comparison between content conditions, our problems are created as \textit{matched sets}. For each condition (except the arbitrary condition), we create problems that are nearly identical in content except for the feature at issue and minor grammatical corrections. We evaluate three different types of problem content:  \textbf{realistic}, \textbf{shuffled}, and \textbf{arbitrary}. A complete diagram of the different problems we evaluate is in Figure ~\ref{fig: design}. Example problems for each condition are in Appendix~\ref{app:ex_problems}.

For the \textbf{realistic} condition, we evaluate a total 140 problems. Of these 140, 70 take the form of \textbf{social rules} and 70 take the form of \textbf{non-social rules}. Of the social rule problems, 35 problems take the form of \textbf{familiar social rules}. These problems are designed to take the form of social rules governing human behavior and are designed to be \textit{familiar} such that they reflect social rules that are consistent with the real world. The other 35 social rule problems take the form of \textbf{unfamiliar social rules}. These problems have the form of social rules but do not contain familiar relationships.

Of the 70 non-social rule problems, 35 are designed to be \textbf{familiar non-social rules} and 35 are designed to be \textbf{unfamiliar non-social rules}. For the \textbf{familiar non-social rule} condition, we evaluate 35 problems that are not social rules and are familiar such that the antecedent and the consequent have a relationship that is consistent with real-world expectations. For the \textbf{unfamiliar non-social rule} condition, we evaluate 35 problems that do not take the form of social rules and are designed such that the antecedent and the consequent do not have a familiar real-world relationship.

The \textbf{realistic} grouping is intended to capture the same types of realistic problems that have been used in previous work \cite{dasgupta_language_2022}. For some of our analyses, we compare these problems as a group.

For the \textbf{shuffled} condition, we evaluate problems where the antecedent and the consequent are switched. We create shuffled versions of each of the different categories of realistic problems. The purpose of the shuffled condition is twofold. First, the creation of shuffled non-social rules allows for the ability to stress the semantics of plausibility beyond mere co-occurrence. Shuffled rules allows us to directly evaluate whether models are sensitive to the words in a problem or the intended semantic meaning. Shuffled problems contain the same words, but convey different conditional logic relationships. Second, the creation of shuffled social rules allows for evaluating sensitivity to the cost-benefit structure of social rules. Standard social rules typically have an if-then format such that if a person receives a benefit, then they must pay the (metaphorical) cost for that benefit, per \cite{cosmides_logic_1989}. In comparison, switched social rules occur in past tense- if a person has paid the cost, then they may receive the benefit. We create shuffled prompts for both the familiar non-social rule and the familiar social rule conditions. We make syntactic corrections to make these problems grammatically correct.

For the \textbf{arbitrary} condition, we evaluate 70 problems where there is no particular relationship between the antecedent and the consequent. For example, in the problem \textit{The rule is that if the cards have a type of food then they must have an outdoor activity}, there is no particular pre-supposed relationship between types of food and outdoor activities.

\subsection{Task Format}

A complete prompt for each problem consists of the instruction sentence, a context sentence, the rule, and a list of cards formatted as a multiple choice question. The instruction sentence was the same for all questions. The context sentences were consistent with the content type. For the arbitrary problems, a neutral context sentence was used to prevent a potential length confounds. The instruction sentence and a sample context sentence are in Appendix~\ref{app:ex_problems}.

\subsection{Models}

For our main set of analyses, we evaluate four recently released large language models with approximately 7 billion parameters: Guanaco, MPT, BLOOM, and Falcon. Guanaco is a family of LLMs fine-tuned with QLoRa, a fine tuning approach designed to reduce memory demands while preserving model performance \cite{dettmers_qlora_2023}. We use the 16-bit version. MPT is an open-source family of LLMs released by Mosaic that are designed to support fast inference \cite{MosaicML2023Introducing, dao_flashattention_2022}. BLOOM was trained on a large multilingual corpus \cite{laurencon_bigscience_2022} and has a decoder-only transformer architecture \cite{workshop_bloom_2023}. Falcon is an open source LLM designed to support fast inference \cite{falcon40b, dao_flashattention_2022, shazeer_fast_2019}. We also run several additional LLMs of varying sizes on this task. Results for these models are in Appendix~\ref{app:large_models}. These models include: the 7B and 13B versions of llama2 \cite{touvron2023llama}, the 7B, 13B, and 30B versions of Wizard \cite{xu_wizardlm_2023}, the 40B version of Falcon \cite{falcon40b}, and the 13B and 33B versions of Guanaco \cite{dettmers_qlora_2023}. See the cited papers for updated details about model licenses.

\subsection{Implementation Details}

We run our experiments on a A100 GPU with 12 vCPUs and 85GB of RAM, running Debian 10. Experiments were conducted in Python 3.10. A complete list of libraries is available in the supplementary materials. Total run time for three main experiments was approximately 3 hours.

\subsection{Evaluation Metrics}
Previous work has proposed various methods to correct for interactions between the specific form of a prompt and the answer generated by a language model \cite{brown_language_2020, holtzman-etal-2021-surface, zhou-etal-2019-unsupervised}. We use one of these metrics, Domain Conditional PMI, as our scoring metric \cite{holtzman-etal-2021-surface}. DCPMI measures how much information a particular instruction domain provides about a particular answer. Formally, a correct answer is equivalent to $$argmax\frac{P(y_i|x)}{P(y_i|baseline)}$$ where $y_i$ is the $i$th answer choice, $x$ is the input prompt, and $baseline$ is the probability associated with a task-specific premise. Candidate answers are evaluated independently. Chance performance is $1/6$. Tables with both traditional accuracy metrics and DCMPI scores for all models can be found in Appendix \ref{app:large_models}.

To facilitate comparison between content conditions, our problems are created as \textit{matched sets}. We model this shared variance statistically via random effects terms for sets of stimuli. We use a mixed-effects approach, which allows for modeling the hierarchical structure of the data \cite{Gelman2006b}. Mixed-effects models are commonly used to analyze linguistic data (i.e., \cite{Matuschek2017, Baayen2008}) and 
permit the generalization of performance beyond a specific set of problems \cite{Clark1973}. We perform follow-up tests by calculating estimated marginal means derived from the entire statistical model for each corresponding analysis. Because estimated marginal means account for other variables in the statistical model, interaction terms may have slightly different coefficients in different analyses. See \cite{lenth_leastsquares_2016} and \cite{searle_population_1980} for additional information.

\section{Results}

\subsection{Analysis 1 Results}

\begin{figure}[t]
\centering

\includegraphics[width=\columnwidth]{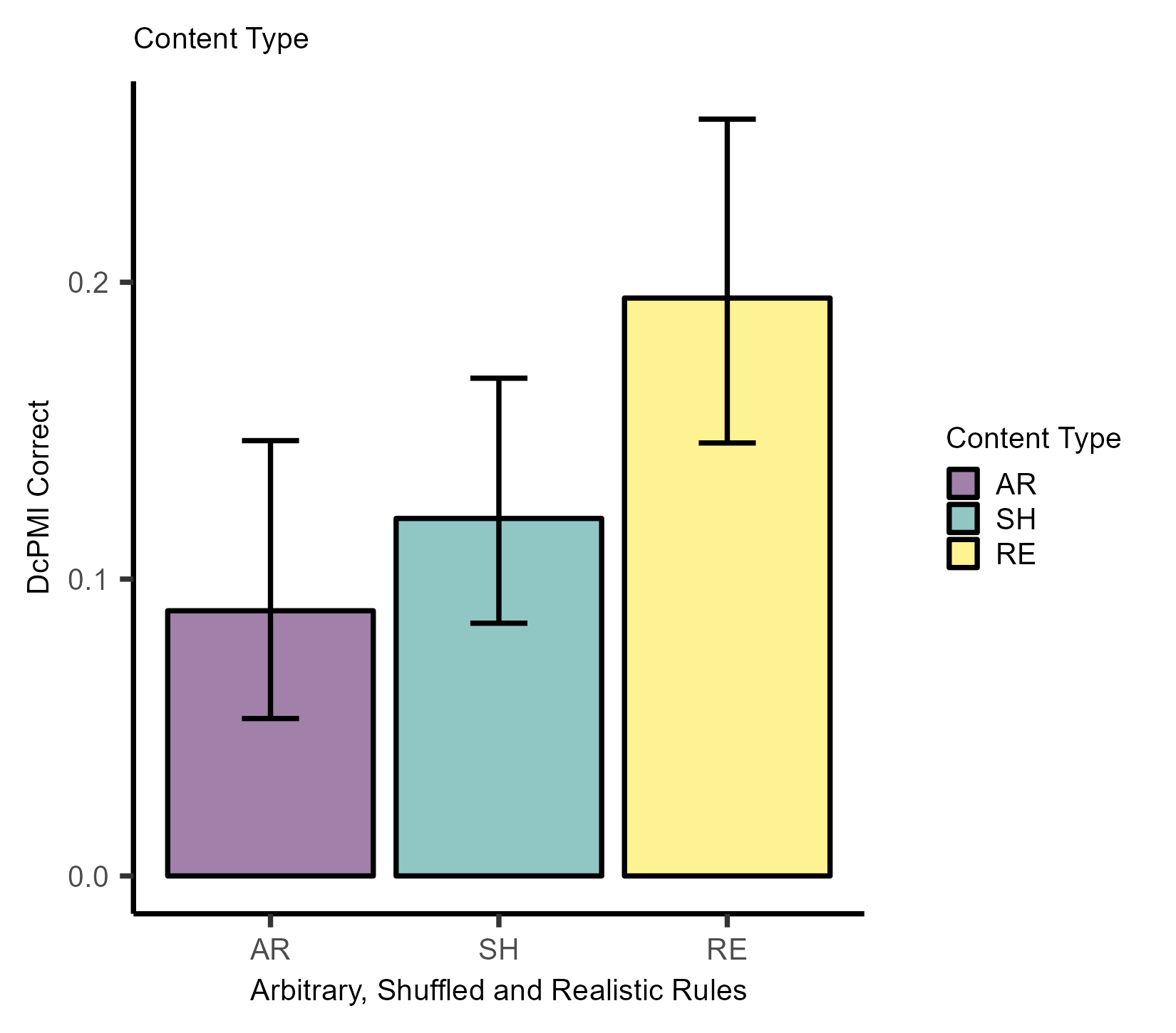}
\caption{Model performance by content type for \textbf{arbitrary} (AR), \textbf{shuffled} (SH), and \textbf{realistic} (RE) rules. RE contains both social and non-social rules. Error bars represent 95 \% confidence intervals. We do not find effects for LLM or familiarity, thus performance is collapsed. Relative to arbitrary content, most models result in a benefit for realistic rules, with mixed influences of shuffling. }
\label{a1overall}
\end{figure}

\begin{figure}[t]
\centering
\includegraphics[width=\columnwidth]{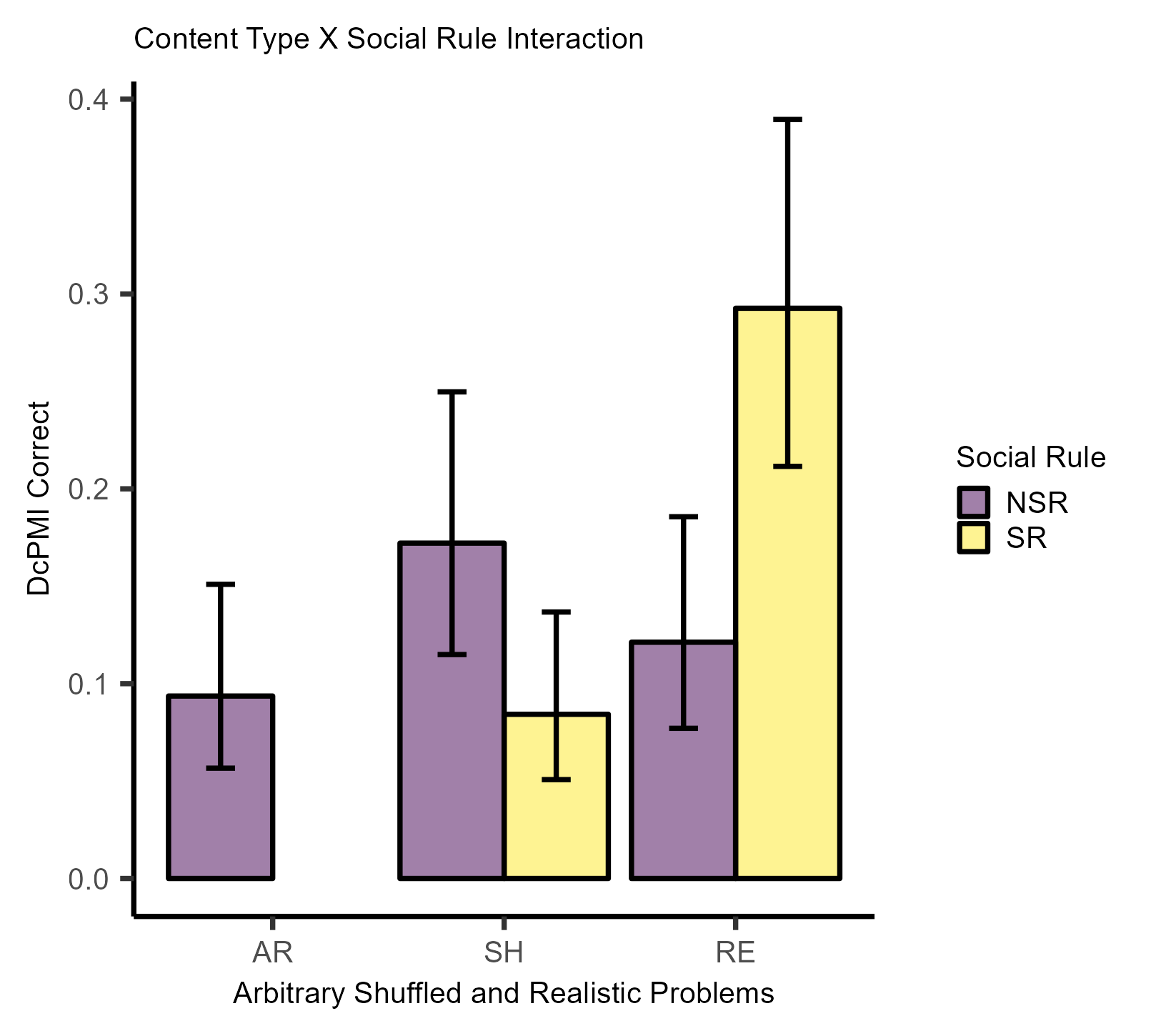} 

\caption{Interaction between content type and social rule status for Analysis 1. Content type: \textbf{arbitrary} (AR), \textbf{shuffled} (SH), or \textbf{realistic} (RE) rules. The realistic category contains social rules and non-social rules. Social rule status: \textbf{social rule} (SR) or \textbf{non- social rule} (NSR) problems. We do not find effects for LLM or familiarity, thus performance is collapsed.}
\label{a1interaction}
\end{figure}

For our first analysis, we evaluate two different crossed factors: social rule status and content type, using DCPMI as our scoring metric. For content type, we evaluate \textbf{arbitrary}, \textbf{shuffled}, and \textbf{realistic} rules. The realistic group contains social rules and non-social rules. We find a significant beneficial main effect for realistic rules compared to shuffled rules and a significant interaction between social rule status and content type (Lines 1 and 2 in Table ~\ref{tab:analysis1} respectively). Factors for LLM and familiarity do not improve model fit, suggesting that the overall pattern of results does not significantly differ between LLMs or between familiar and unfamiliar problems. See Appendix~\ref{app:follow-up} for follow-up interaction tests. Overall performance is illustrated in Figure \ref{a1overall}; see Figure \ref{a1interaction} for the interaction. Performance remains rather low overall.

\begin{table}[]
    \centering
    \begin{tabular}{p{0.39\columnwidth}p{0.5\columnwidth}}
    \hline
    \textbf{Effect} & \textbf{OR, CI, $Z$} \\
    \hline
    RE v SH & 1.30 [1.05 - 1.62] 2.29*\\
    SR Status x Content  & 1.59 [1.26 - 2.02] 3.97**\\ 
    \hline
    RE NSR v RE SR & 0.33 [0.22 - 0.44] -3.34**\\
    SH NSR v SH SR & 2.25 [1.47 - 3.03] 2.35* \\
    SH SR v RE SR & 0.22 [0.15 - 0.29] -4.38** \\
    \hline
    \end{tabular}
    \caption{Statistical results for analysis 1. AR=arbitrary, SH=shuffled, RE=realistic, SR=social rule, NSR=non-social rule, OR=odds ratio, CI=confidence interval, * = $p<0.05$, ** = $p<0.01$. Interactions in bottom half.}
    \label{tab:analysis1}
\end{table}

\begin{figure}[t]
\centering
\includegraphics[width=\columnwidth]{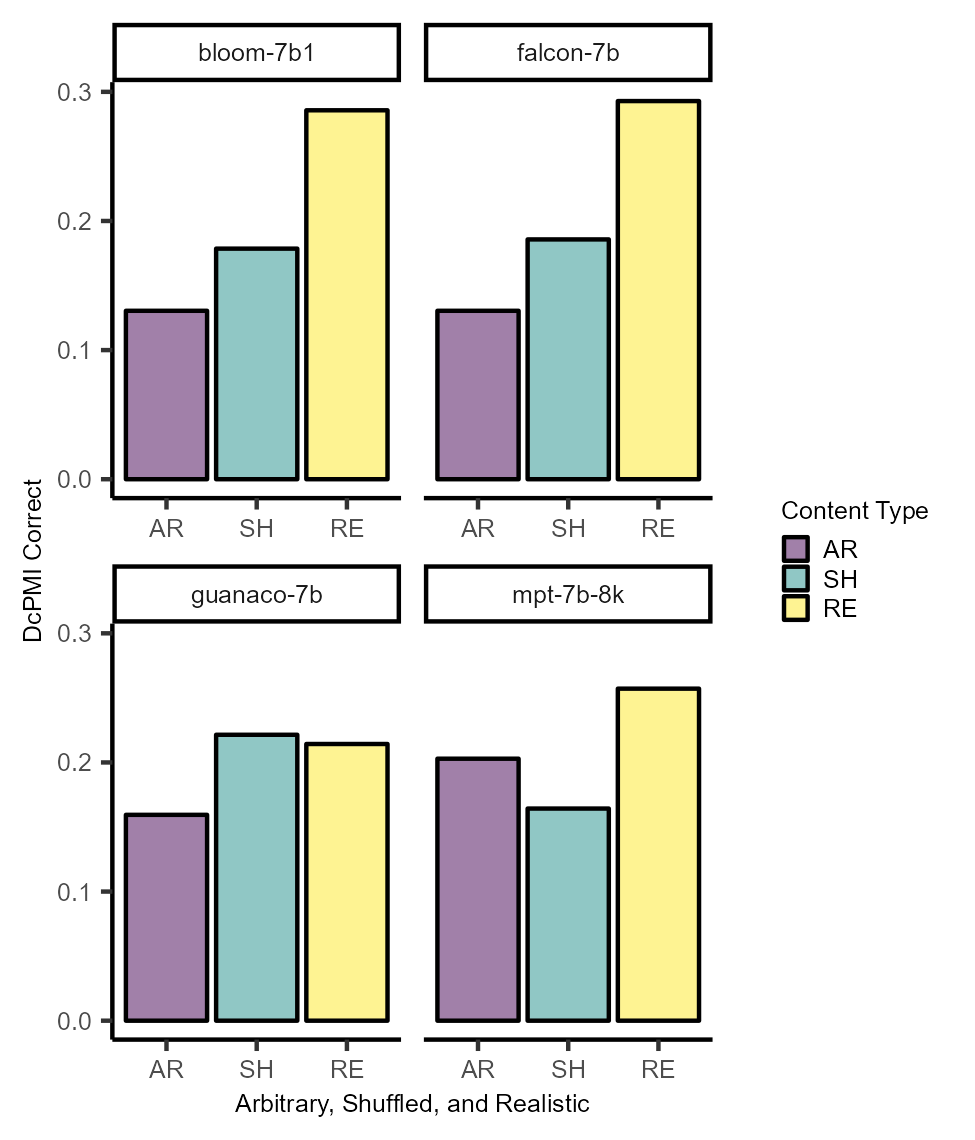}

\caption{Performance across all models for \textbf{arbitrary} (AR), \textbf{shuffled} (SH), and \textbf{realistic} (RE) rules. The realistic category contains both social rules and non-social rules. We collapse across LLM and familiarity.}
\label{a1bymodel}
\end{figure}

\subsection{Analysis 1 Discussion}

Initial results from analysis 1 do seem to replicate the general \textit{pattern} of results demonstrated for human subjects: performance is better for social rule problems than non-social rule problems. We also find an effect for switched social rules such that switched social rules have considerably lower performance than standard social rule problems. This effect is similar to one reported for human subjects  \cite{cosmides_logic_1989}. 

However, we do not replicate the \textit{magnitude} of the content effect. While humans find abstract problems quite difficult, they successfully solve social rule problems \~70\% of the time \cite{griggs_elusive_1982}. In comparison, the LLMs solve social rule problems approximately 30\% of the time when using DCPMI scoring. Comparison with traditional highest probability scoring indicates that this pattern of results is not due to domain conditional scoring; highest probability scoring produces worse overall performance and the overall pattern of results is similar.

Additionally, we find an interaction between social rule status and content type. This interaction demonstrates that LLMs are sensitive to some aspects of the structure of social rules. This is somewhat consistent with results from human subjects which predict that performance should be lower for shuffled social rules compared to standard social rules \cite{cosmides_logic_1989}. However, human subject responses do not predict the observed differences between shuffled social rules and shuffled non-social rules. Performance for LLMs is influenced by problem content, but in a manner that is not parallel to human behavior.

\subsection{Analysis 2 Results}
A follow-up experiment examines whether alternative presentation formats may improve performance, given some previous results that suggest LLMs may have better reasoning performance with more explicit representations (e.g., \citealp{saparov_testing_2023}). In addition to the standard presentation format (\textbf{classic}), we investigate three additional formats. In the \textbf{front} condition, the problems include descriptions of the front of each card. In the \textbf{back} condition, the problems include descriptions of the hypothetical category of the item on the back of the card. In the \textbf{both} condition, the problems include descriptions of both the front and the hypothetical category on the back of the card. We created alternative formats for all of the content types: \textbf{realistic social rules}, \textbf{realistic non-social rules}, \textbf{shuffled social rules}, and \textbf{shuffled non-social rules}. We use DCPMI as our scoring metric. 

The best fitting statistical model includes an interaction between presentation format, social rule status, and content type plus a random effect for item instances. Adding factors for LLM and problem familiarity did not improve overall model fit, suggesting that performance does not vary substantially by model or by familiarity. 

The main effect for presentation format was significant. Comparisons between classic versus front, front versus back, and back versus both presentation formats were all significant (lines 1, 2, and 3 respectively in the top half of Table~\ref{tab:analysis2}). The main effects for social rule status and content type were not significant.

For the two-way interaction between social rule status and presentation format, comparisons between social rule status and front versus back presentation formats and social rule status and back versus both presentation formats were significant (lines 1 and 2 respectively in the bottom half of Table~\ref{tab:analysis2}).

For the two-way interaction between content type and presentation format, the comparison between shuffled versus realistic content types and classic and front presentation formats and the comparison between shuffled versus realistic content types and front versus back presentation formats were significant (lines 3 and 4 respectively in the bottom half of Table~\ref{tab:analysis2}). 

The two-way interaction between social rule status and content type (shuffled or realistic) was significant (lines 5 in the bottom half of Table~\ref{tab:analysis2}). 

\begin{figure}[t]
\centering
\includegraphics[width=\columnwidth]{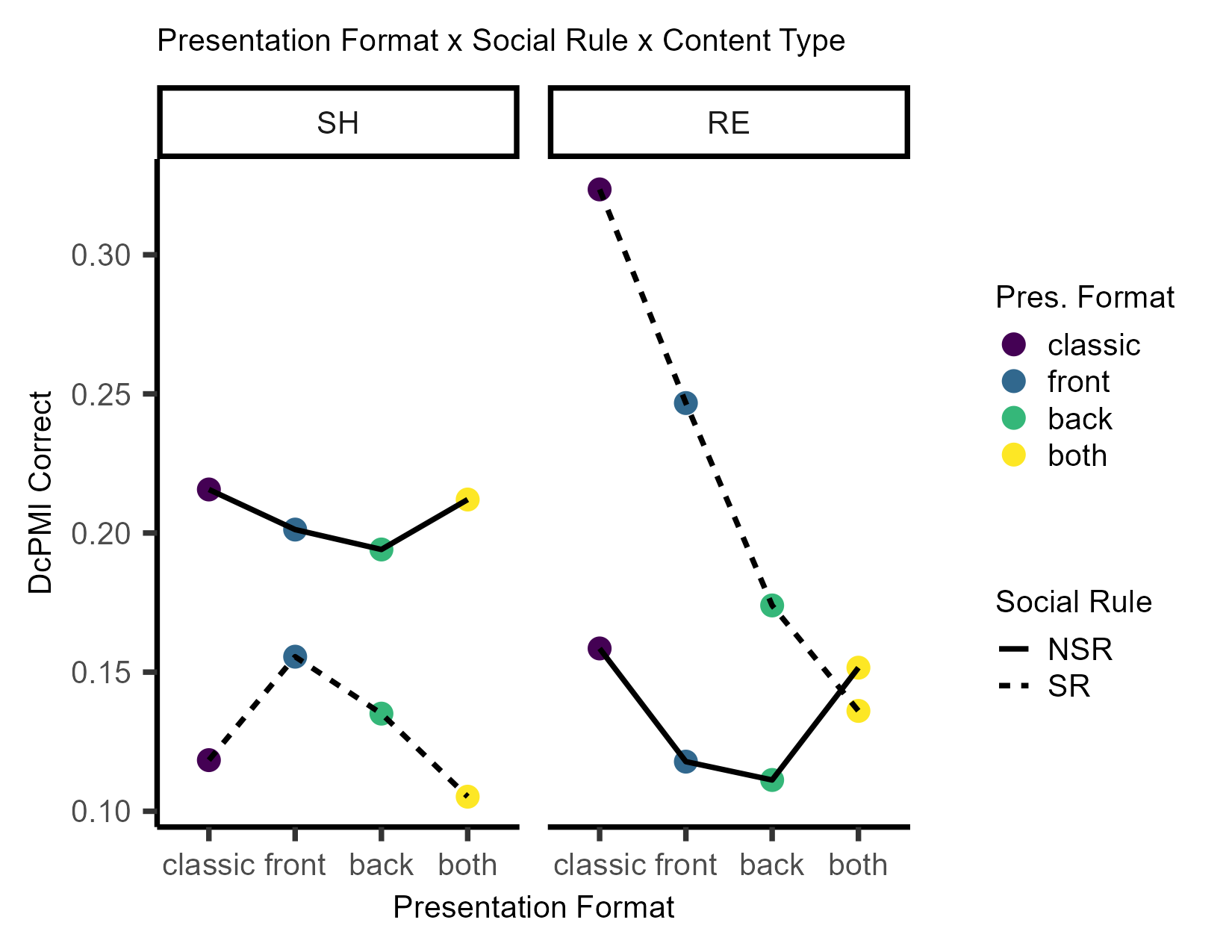}

\caption{Interaction between presentation format (\textbf{classic}, \textbf{front}, \textbf{back}, or \textbf{both}), content type (\textbf{shuffled} (SH) or \textbf{realistic} (RE)), and social rule status (\textbf{social rule} or \textbf{non social rule}) broken out by presentation format.}
\label{a2interaction1}
\end{figure}

\begin{figure}[t]
\centering
\includegraphics[width=\columnwidth]{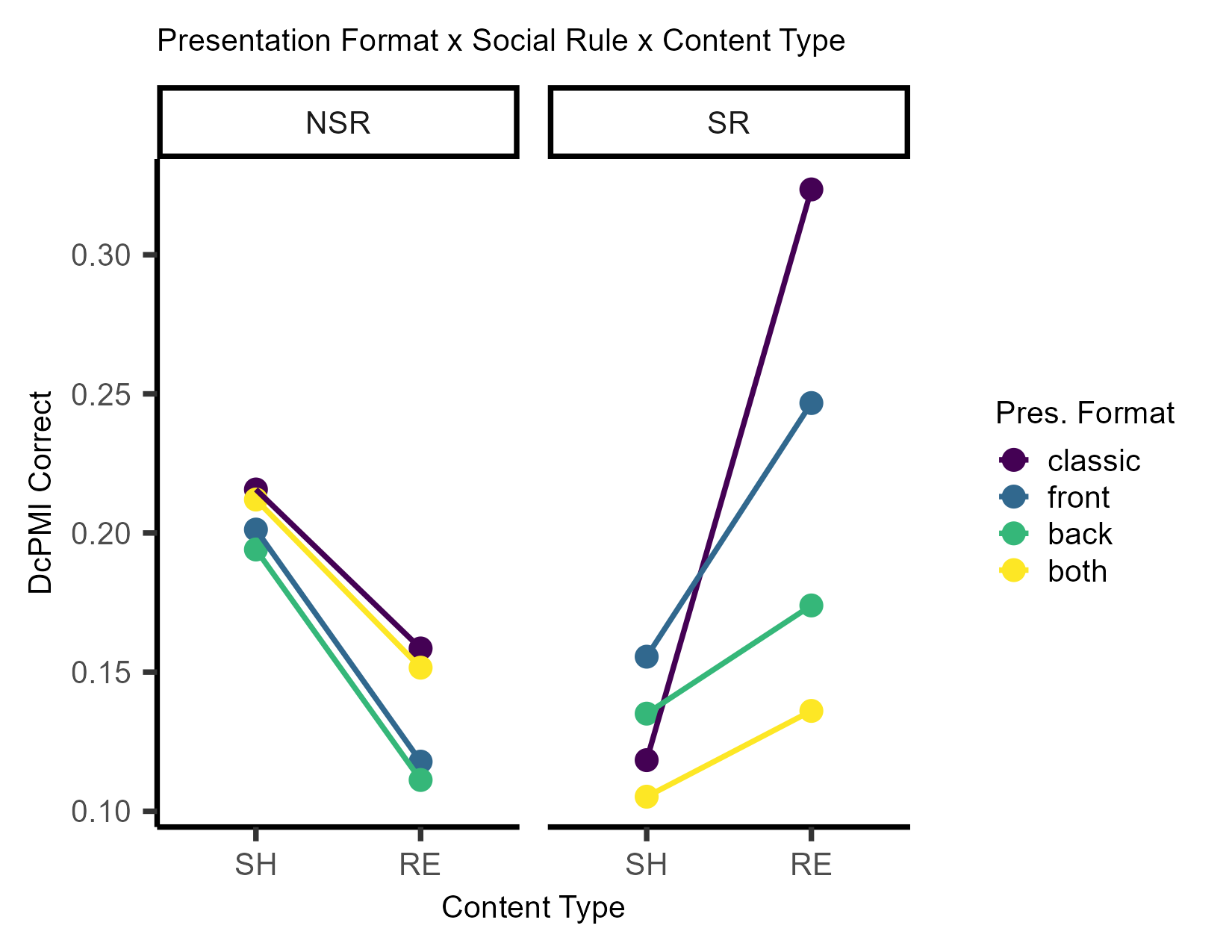} 

\caption{Interaction between presentation format (\textbf{classic}, \textbf{front}, \textbf{back}, or \textbf{both}), content type (\textbf{shuffled} (SH) or \textbf{realistic} (RE)), and social rule status (\textbf{social rule} or \textbf{non social rule}) broken out by content type.}
\label{a2interaction2}
\end{figure}

No three way interactions were significant ($z$-values $= (1.92,1.88,1.50)$, all $p>0.05$). See Figures ~\ref{a2interaction1} and ~\ref{a2interaction2} for plots of the interactions. Follow up tests for individual level comparisons within each two-way interaction are located in Appendix~\ref{app:follow-up2}.

Problem familiarity did not improve model fit, suggesting that performance does not vary according to the familiarity of problem content.

\begin{table}
\centering
\begin{tabular}{p{0.415\columnwidth}p{0.48\columnwidth}}
\hline
\textbf{Main Effect} & \textbf{OR, CI, $Z$} \\
\hline
Classic v Front & 1.19 [1.06 - 1.34] 2.79**\\
Front v Back  & 1.28 [1.09 - 1.50] 3.17**\\ 
Back v Both & 1.15 [1.00 - 1.31] 1.98*\\
\hline
SR v NSR x F v B & 1.22 [1.06 - 1.40] 2.6*\\
SR v NSR x B v Both & 1.25 [1.09 - 1.44] 3.30**\\ 
SH v RE x C v F & 1.22 [1.08 - 1.37] 2.96** \\
SH v RE x F v B & 1.17 [1.02 - 1.34] 2.04* \\
SR v NSR x SH v RE & 1.32 [1.17 - 1.48] 4.66** 
\\
\hline
\end{tabular}

\caption{Statistical results for analysis 2. AR=arbitrary, SH=shuffled, RE=realistic, SR=social rule, NSR=non-social rule, OR=odds ratio, CI=confidence interval, * = $p<0.05$, ** = $p<0.01$. Interactions in bottom half.}
\label{tab:analysis2}
\end{table}

\subsection{Analysis 2 Discussion}
In general, we find that models are sensitive to different presentation formats. However, we do not find performance improvements for different presentation formats. Results suggest that there are some interactions between presentation format and problem content; however, most of our follow up tests were not significant. A follow up analysis of treatment effects broken out by content type suggests that presentation formats have more of an effect on realistic rules than shuffled rules. Such interactions are not anticipated in human performance data (e.g., \citealp{wason_reasoning_1984, manktelow_facilitation_1979}).

\subsection{Analysis 3 Results}
For analysis 3, we examine antecedent selection, also scored with 
DCPMI, across all conditions. Specifically, we evaluate whether content type, presentation format, or social rule status influences whether the models select answers that contain any antecedent card. Complete statistical results are in Table ~\ref{tab:analysis3}.

The best-fitting statistical model contains a main effect for social rule status and an interaction between presentation format and content type, plus a random effect for overall item. A term for LLM did not improve model fit. For the presentation formats main effect, we find significant main effects for classic versus front presentation formats, front versus back presentation formats, and back versus both presentation formats (lines 1, 2, and 3 respectively in Table~\ref{tab:analysis3}). For the content type main effect, we find significant differences between arbitrary versus non-arbitrary problems and between shuffled and realistic problems (lines 4 and 5 respectively in Table ~\ref{tab:analysis3}). The main effect for social rule status was not significant. 

For the interaction between presentation format and content type, we find significant differences for contrasts between arbitrary and non-arbitrary problems and classic versus front presentation formats, shuffled versus realistic problems and classic versus front presentation formats, shuffled versus realistic problems and front versus back formats, shuffled versus realistic problems and back versus both formats (Lines 1-4 respectively in Table ~\ref{tab:analysis3_maineffect_interaction}). Interactions are displayed in Figure~\ref{a3I}. Follow-up tests for each of the interaction effects can be found in Table~\ref{tab:analysis3app} in Appendix \ref{app:follow-up}.

\begin{table}
\centering
\begin{tabular}{ll}
\hline
\textbf{Effect} & \textbf{OR, CI, $Z$} \\
\hline
C v F & 1.34 [1.15 - 1.57] 3.76**\\
F v B  & 1.29 [1.10 - 1.51] 3.10**\\
B v Both & 1.23 [1.07 - 1.41] 3.06*\\
AR v NAR & 1.53 [1.23 - 1.90] 3.74**\\
SH v RE & 1.30 [1.14 - 1.50] 3.59**\\

\hline
\end{tabular}

\caption{Statistical results for antecedent selection in analysis 3. AR=arbitrary, NAR=non-arbitrary, SH=shuffled, RE=realistic, SR=social rule, NSR=non-social rule, C=classic, F=front, B=back, OR=odds ratio, CI=confidence interval, * = $p<0.05$, ** = $p<0.01$}
\label{tab:analysis3}
\end{table}

\begin{figure}[t]
\centering
\includegraphics[width=\columnwidth]{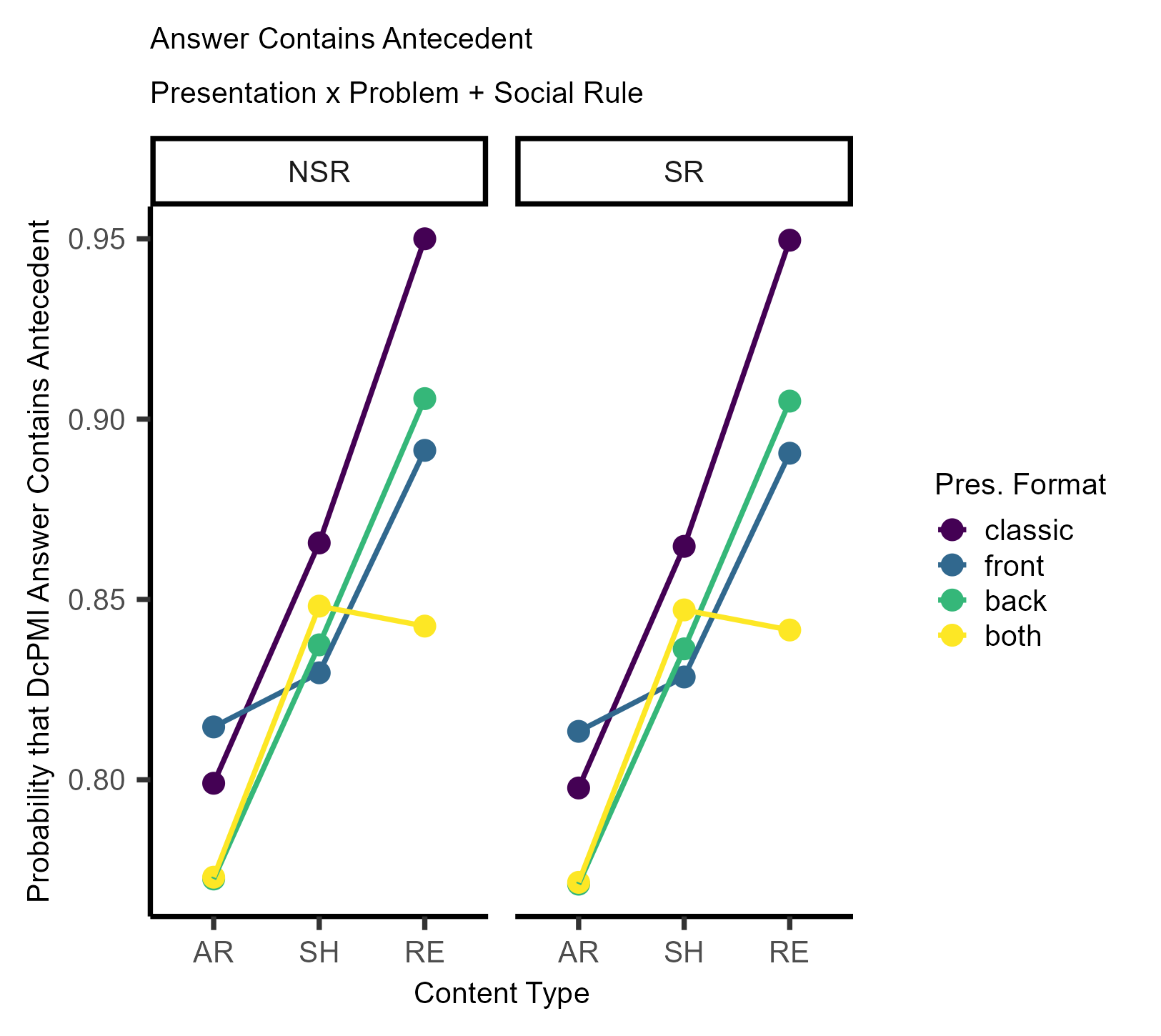} 

\caption{Evaluation of whether the LLMs select an antecedent card. Content type: \textbf{arbitrary} (AR), \textbf{shuffled} (SH), and \textbf{familiar} (FM). Presentation formats: \textbf{classic}, \textbf{front}, \textbf{back}, and \textbf{both}. Social rule status: \textbf{non-social rule}, \textbf{social rule}. Collapsed over LLM.}
\label{a3I}
\end{figure}

\subsection{Analysis 3 Discussion}

Results from analysis 3 are particularly interesting because there is limited variance in human performance according to this metric. Regardless of content type, human subjects select at least one antecedent card \cite{johnson-laird_reasoning_1972, griggs_elusive_1982}. Even for conditions that influence antecedent card selection, the overwhelming tendency is for participants to select the alternative antecedent card \cite{cosmides_logic_1989}. In contrast, our results suggest that whether LLMs select antecedent cards varies significantly according to content type and presentation format. 

Despite differences in training datasets and tasks and model architecture, we do not find any effects for LLM. Additionally, the interaction between content type and presentation format suggests that different presentation formats have differential effects on antecedent selection for different content types. LLMs may benefit from different types of formatting depending on the content of the reasoning task.

\section{Discussion}
We set out to expand the research base on evaluating the reasoning capabilities of LLMs with a classic experiment contrasting \textit{a priori} conditions.  We do replicate some effects found in the literature for human subjects. Performance is higher for familiar problems than arbitrary ones and social rule have higher performance than non-social rules. However, we do not replicate the magnitude of the effects; social content does not benefit LLM performance as might be expected based on training corpus content. 

We do find effects for different presentation formats; however, they do not improve performance. For shuffled problems, the specific presentation format does not make a difference. For realistic problems, we find presentation format does make a difference-the classic presentation format has the highest performance. 

However, our systematic, content and format controlled experimentation and performance measurement has also revealed a number of inexplicable interactions that \textit{appear to be consistent across different LLMs} and are therefore independent of architecture. Given the literature on human performance, an interaction between social rule status and content type is expected. However, many of the other interactions are not expected and are inconsistent with human performance (e.g., \citealp{wason_reasoning_1984, manktelow_facilitation_1979}).

We find some evidence that LLMs benefit from different types of presentation formats, (depending on the specific content of the problem), as might be expected from popular compensatory prompt engineering efforts. However, it is not immediately clear what types of information facilitate overall reasoning performance. This limits the ability to make general predictions about the conditions under which LLM reasoning is accurate. 

In addition to content and presentation interactions, we find that LLMs do not pick antecedent cards at the same rate that human subjects do. Moreover, this behavior is influenced by the task condition- LLMs are less likely to select antecedent cards for arbitrary and realistic problems than for social rules.

Overall low performance is particularly surprising for realistic social and non-social rules as the relationships for solving these problems are plausibly available in the training data of LLMs.  Yet we find that all LLMs diverge from documented human performance. 

Overall performance is also remarkably consistent across models despite different training data, objectives, and model architectures. Moreover, we find that the interaction results are also independent of the LLM examined. Some consistency between LLMs is to be expected, given that LLMs are trained on human-generated text corpora. However, the commonalities are not consistent with human behavior. This suggests a common yet surprising emergent reasoning bias without any apparent adaptive benefit. 

Fine-tuning the models for this task would likely improve task performance. We did not fine-tune the models for several reasons. First, the Wason task is intended to be a general task for evaluating reasoning performance. These tap into general knowledge and a set of reasoning skills that transfers to new tasks. Thus, the position that networks specifically trained on large reasoning task corpora is the best way to evaluate the reasoning performance of models is questionable. This is particularly true given that models often have high performance on the specific training task and limited performance on related reasoning tasks \cite{mitchell_abstraction_2021}.

Second, several researchers have proposed that the Wason task can be solved via linguistic and real world knowledge \cite{pollard_human_1982,tversky_availability_1973, wason_realism_1983}. Human participants achieve high performance on problems that deal with familiar social rules with no experience with the task, using prior knowledge. 

However, the knowledge that this task requires is plausibly available in the training data for LLMs. The words used in this task are all common English words. Moreover, many of the relationships between items are plausibly available in training text, particularly for problems that deal with familiar social rules. Yet, performance remained quite low. 

Previous work has proposed that ideal tasks for evaluating the reasoning of computational algorithms are those that do not require task-specific training \cite{chollet_measure_2019, mitchell_abstraction_2021}. We concur with this position and suggest that the Wason task is an ideal task in this regard.

\section{Conclusion}
Despite substantial performance improvements on standard benchmark datasets, existing LLMs have considerable room for improvement with regards to many aspects of human intelligence \cite{lake_building_2017, mitchell_abstraction_2021}. In these experiments, we specifically investigate two of these aspects: generalized performance on related tasks and generation of answers at the limits of available knowledge. 

Overall, our results replicate some of the same patterns found in the cognitive science literature. However, performance remains poor with inexplicable interactions between problem content and our efforts to manipulate presentation format. LLMs are sensitive to different sets of task criteria than human subjects. These criteria are not predictable across conditions and suggest areas where the reasoning of LLMs is not consistent with that of human capability.

\bibliography{new_anthology,zotero}
\bibliographystyle{acl_natbib}

\appendix

\section{Example Problems}
\label{app:ex_problems}

In this section, we provide example problems. For details on the conditions and their associated rationale, see Section~\ref{sec:introduction}.

\subsection{Definitions}

\begin{figure}[t]
    \centering
    \includegraphics{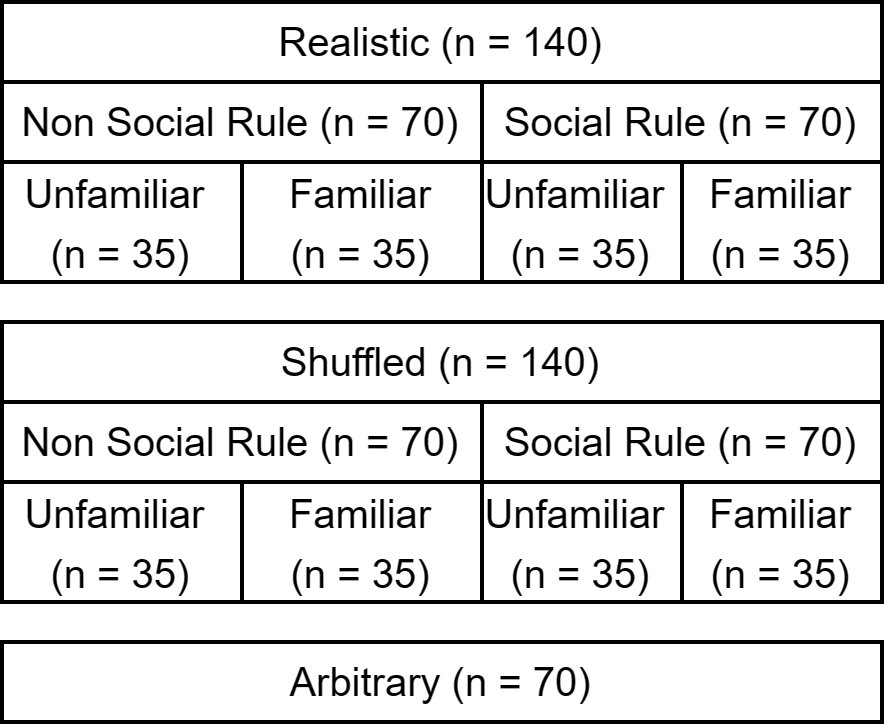}
    \caption{Breakdown of the different types of problems we examine.}
    \label{fig: design}
\end{figure}

Below are definitions and examples of the different problem components. A complete design matrix for the study is provided in ~\ref{fig: design}.

\textbf{Context Sentence:}  Used as context for for the problem. Example: \textit{An attendant needs to make sure that customers are following the rules.} \\
\textbf{Instruction sentence}: Used at the end of all problems to prompt the model. Example: \textit{Pick two cards that are required to determine if the rule is true.} \\
\textbf{Familiar Social Rule}: Problems that take the form of a social rule (i.e., \textit{If a person pays the cost, they receive the benefit}) that deals with familiar relationships. Example: \textit{The rule is that if the customer is over 25 they can drive a rental car.} \\
\textbf{Unfamiliar Social Rule}: Problems that take the form of a social rule (i.e., \textit{If a person pays the cost, they receive the benefit}) and deals with unfamiliar relationships. Example: \textit{The rule is that if the customer is over 25 they must be in elementary school.} \\
\textbf{Familiar Non-social Rule}: Problems that do not take the form of a social rule and deal with familiar relationships. Example: \textit{The rule is that if the equipment is a laptop then it must have a plastic keyboard}. \\
\textbf{Unfamiliar Non-social Rule}: Problems that do not take the form of a social rule and deal with unfamiliar relationships. Example: \textit{The rule is that if the equipment is a laptop then it must have a grass keyboard}. \\
\textbf{Shuffled Rules:} Created from the rule types above. Shuffling allows us to evaluate the extent to which models are sensitive to the cost benefit structure of the rules. Example \textbf{shuffled familiar social rule}: \textit{The rule is that if the customer can drive a rental car they must be over 25.}
\textbf{Arbitrary}: Rules that contain arbitrary relationships. Example: \textit{The rule is that if the cards have a type of food then they must have an outdoor activity.} \\

\subsection{Complete Example Problems}

Below are complete example problems for each of the presentation conditions. Note that the order of the answers was shuffled in the actual experiments, answers are presented in here the $P, \neg P, Q, \neg Q$ order for all problems for ease of interpretation.

\textbf{Familiar Social Rule (Classic)}: \textit{Context:} Pick two cards that are required to determine if the rule is true: \textit{Problem:} An attendant needs to make sure that customers are following the rules. The rule is that if the customer is over 25 they can drive a rental car: A: 27 years old B: 15 years old C: driving rental car D: riding the bus

\textbf{Familiar Social Rule (Front)}: \textit{Context:} Pick two cards that are required to determine if the rule is true: \textit{Problem:} An attendant needs to make sure that customers are following the rules. The rule is that if the customer is over 25 they can drive a rental car. The first card has 15 years old on the front. The second card has driving rental car on the front. The third card has riding the bus on the front. The fourth card has 27 years old on the front: A: 27 years old B: 15 years old C: driving rental car D: riding the bus

\textbf{Familiar Social Rule (Back)}: \textit{Context:} Pick two cards that are required to determine if the rule is true: \textit{Problem:} An attendant needs to make sure that customers are following the rules. The rule is that if the customer is over 25 they can drive a rental car. The first card has an unknown age on the back. The second card has an unknown activity on the back. The third card has an unknown age on the back. The fourth card has an unknown activity on the back: A: 27 years old B: 15 years old C: driving rental car D: riding the bus

\textbf{Familiar Social Rule (Both)}: \textit{Context:} Pick two cards that are required to determine if the rule is true: \textit{Problem:} An attendant needs to make sure that customers are following the rules. The rule is that if the customer is over 25 they can drive a rental car. The first card has riding the bus on the front and an unknown age on the back. The second card has driving rental car on the front and an unknown age on the back. The third card has 15 years old on the front and an unknown activity on the back. The fourth card has 27 years old on the front and an unknown activity on the back: A: 27 years old B: 15 years old C: driving rental car D: riding the bus

\section{Follow-up Interaction Tests}
\label{app:follow-up}

\subsection{Analysis 1 Follow-up Tests}
Results from analysis 1 found support for one interaction between social rule status and content type.

\subsubsection{Content Type and Social Rule Status}

Three of the follow-up tests for the interaction between social rule status and content type were significant. The test comparing realistic non-social rules and realistic social rules test was significant (Bottom half of Table ~\ref{tab:analysis1}, line 1). The test comparing shuffled non-social rules and shuffled social rules was significant (Bottom half of Table ~\ref{tab:analysis1}, line 2). The test comparing shuffled social rules and realistic social rules was significant (Bottom half of Table ~\ref{tab:analysis1}, line 3).  

\begin{table}
    \centering
    \begin{tabular}{ll}
        \hline
        \textbf{Interaction Effect} & \textbf{OR, CI, $Z$} \\
        \hline 
        AR v NAR x C v F & 1.28 [1.03 - 1.59] 2.17*\\
        SH v RE x C v F & 1.29 [1.08 - 1.54] 2.74**\\
        SH v RE x F v B & 1.28 [1.05 - 1.56] 2.54*\\
        SH v RE x B v Both & 1.33 [1.14-1.56] 3.61** \\
        \hline
    \end{tabular}
    \caption{Effects for interactions for Analysis 3. AR=arbitrary, NAR=non-arbitrary, SH=shuffled, RE=realistic, SR=social rule, NSR=non-social rule, C=classic, F=front, B=back, OR=odds ratio, CI=confidence interval, * = $p<0.05$, ** = $p<0.01$}
    \label{tab:analysis3_maineffect_interaction}
\end{table}

\subsection{Analysis 2 Follow-up Tests}
\label{app:follow-up2}

Results from analysis 2 found support for three two-way interactions: one between social rule status and presentation format, one between content type and presentation format, and one between social rule status and content type. 

\subsubsection{Social Rule Status and Presentation Format}

None of the follow-up tests were significant for the front versus back or the back versus both presentation format interactions with social rule status. 

\subsubsection{Content Type and Presentation Format}

Follow-up tests for the shuffled versus realistic rules and classic versus front presentation format contrast found significant differences between shuffled and realistic rules for the classic problems (line 1 Table~\ref{tab:analysis2app}) and other tests were non-significant. 

All follow-up tests for the shuffled versus realistic rules and the front versus back presentation format contrast were non-significant. 
 
We conducted a follow-up set of comparisons to examine differences within the shuffled and realistic content types. We found significant differences between social rules within the both presentation format (line 4 Table~\ref{tab:analysis2app}). For realistic rules, all four comparisons were significant (lines 5-8 Table~\ref{tab:analysis2app}).

\subsubsection{Social Rule Status and Content Type}

For the interaction between social rule status and content type (shuffled versus realistic), follow-up tests between shuffled vs realistic non-social rules (line 2 Table~\ref{tab:analysis2app}) and realistic social rules vs shuffled social rules were significant (line 3 Table~\ref{tab:analysis2app}).

\begin{table}
\centering
\begin{tabular}{ll}
\hline
\textbf{Interaction Effect} & \textbf{OR, $Z$, CI} \\
\hline
SH C v RE C & 1.56, 2.5*, [1.29-1.83] \\
SH SR v RE NSR & 1.68, 3.04**, [1.39 - 1.96]\\
RE SR v SH SR & 1.84, 3.55**, [1.53 - 2.15]\\

SH SR both & 0.60, -2.78* \\
RE NSR F & 0.65, -2.40* \\
NSR B & 0.61, -2.73* \\
SR C & 2.35, 5.95** \\
SR F & 1.61, 3.20** \\
\hline
\end{tabular}

\caption{Follow-up interaction tests for Analysis 2. AR=arbitrary, NAR=non-arbitrary, SH=shuffled, RE=realistic, SR=social rule, NSR=non-social rule, C=classic, F=front, B=back, OR=odds ratio, CI=confidence interval, * = $p<0.05$, ** = $p<0.01$}
\label{tab:analysis2app}
\end{table}

\subsection{Analysis 3 Follow-up Tests}

\begin{table}
\centering
\begin{tabular}{ll}
\hline
\textbf{Follow-up Test} & \textbf{OR, CI, $Z$} \\
\hline
AR C v NAR C & 2.73 [1.49-5.19] 4.09**\\
NAR C v NAR F & 0.57 [0.39-0.83] -3.68**\\

SH C v RE C & 0.33 [0.17-0.65] -4.091**\\
SH F v RE F & 0.59 [0.34-1.0] -2.35*\\
SH F v RE C & 0.25 [0.13-0.49] -5.23*\\

SH F v RE F & 0.59 [0.34-1.0] -2.35*\\
SH B v RE B & 0.53 [0.30-0.93] -2.77*\\
SH B v RE F & 0.62 [0.36-1.0] -2.09*\\
SH F v RE B & 0.50 [0.29-0.88] -3.03**\\

SH Back v RE Back & 0.53 [0.30-0.93] -2.77*\\
SH Both v RE Back & 0.58 [0.33-1.0] -2.41*\\

\hline
\end{tabular}

\caption{Follow-up interaction tests for analysis 3. AR=arbitrary, NAR=non-arbitrary, SH=shuffled, RE=realistic, SR=social rule, NSR=non-social rule, C=classic, F=front, B=back, OR=odds ratio, CI=confidence interval, * = $p<0.05$, ** = $p<0.01$}
\label{tab:analysis3app}
\end{table}

Analysis 3 found support for an interaction between content type (arbitrary, shuffled, or realistic) and presentation format (classic, front, back, or both). 

\subsubsection{Content Type and Presentation Format}

Tests for this interaction indicated significant differences for four contrasts: one between arbitrary and non-arbitrary content types and classic versus front presentation formats (1), one between shuffled versus realistic content types and classic versus front presentation formats (2), one between shuffled versus realistic content types and front versus back presentation formats (3), and one between shuffled versus realistic content types and back versus both presentation formats (4).

Contrast 1: For the contrast between arbitrary and non-arbitrary content types and classic versus front presentation formats, two follow-up tests were significant. These include: 1) the test between arbitrary classic versus non-arbitrary classic (line 1, Table ~\ref{tab:analysis3app}) and 2) the test between non-arbitrary classic and non-arbitrary front (line 2, Table ~\ref{tab:analysis3app}). 

Contrast 2: For the contrast between shuffled and realistic content types and classic versus front presentation formats, three of the follow-up tests were significant. These include: the test between shuffled classic and realistic classic (line 3, Table ~\ref{tab:analysis3app}), shuffled front and realistic front (line 4, Table ~\ref{tab:analysis3app}), and shuffled front and realistic classic (line 5, Table ~\ref{tab:analysis3app}).

Contrast 3: For the contrast between shuffled and realistic content types and front versus back presentation formats, all four follow up tests were significant. These include: shuffled front versus realistic front (line 6, Table ~\ref{tab:analysis3app}), shuffled back versus realistic back (line 7, Table ~\ref{tab:analysis3app}), shuffled back versus realistic front (line 8, Table ~\ref{tab:analysis3app}), and shuffled front versus realistic back (line 9, Table ~\ref{tab:analysis3app}).

Contrast 4: For the contrast between shuffled versus realistic content types and back versus both presentation formats, two follow-up tests were significant. These include: the follow up test between shuffled back versus realistic back (line 10, Table ~\ref{tab:analysis3app}) and shuffled both versus realistic back (line 11, Table ~\ref{tab:analysis3app}). Interaction plots are in Figure ~\ref{a3I}.

\section{Additional Models}
\label{app:large_models}

In this section, we report full accuracy (see Table~\ref{tab:additional-models-lm}) and domain-conditional PMI (see Table~\ref{tab:additional-models-dcpmi}) scores for all the models in the main paper, as well as several additional models. While we do find that some of the larger models perform somewhat better, the overall pattern of results is similar.

\begin{table*}
\centering
\begin{tabular}{llllll}
\hline
\textbf{Model} & \textbf{Cond} & \textbf{Classic} &\textbf{Front} & \textbf{Back} & \textbf{Both} \\
\hline

guanaco-7b & AR & 0.1 (0.3) & 0.14 (0.35) & 0.16 (0.37) & 0.09 (0.28)\\
guanaco-7b & SH & 0.1 (0.3) & 0.16 (0.37) & 0.14 (0.34) & 0.14 (0.34)\\
guanaco-7b & SR & 0.06 (0.25) & 0.09 (0.29) & 0.09 (0.29) & 0.07 (0.26)\\

mpt-7b-8k & AR & 0.06 (0.23) & 0.21 (0.41) & 0.09 (0.28) & 0.09 (0.28)\\
mpt-7b-8k & SH & 0.09 (0.28) & 0.14 (0.34) & 0.09 (0.29) & 0.19 (0.39)\\
mpt-7b-8k & SR & 0.14 (0.34) & 0.15 (0.36) & 0.15 (0.36) & 0.11 (0.32)\\

bloom-7b1 & AR & 0.09 (0.28) & 0.24 (0.43) & 0.13 (0.34) & 0.11 (0.32)\\
bloom-7b1 & SH & 0.13 (0.34) & 0.13 (0.34) & 0.07 (0.26) & 0.15 (0.36)\\
bloom-7b1 & SR & 0.16 (0.37) & 0.16 (0.37) & 0.11 (0.32) & 0.08 (0.27)\\

falcon-7b & AR & 0.11 (0.32) & 0.21 (0.41) & 0.06 (0.23) & 0.06 (0.23)\\
falcon-7b & SH & 0.09 (0.28) & 0.11 (0.32) & 0.07 (0.26) & 0.11 (0.32)\\
falcon-7b & SR & 0.14 (0.34) & 0.16 (0.37) & 0.13 (0.34) & 0.11 (0.32)\\

WizardLM-7B-V1.0 & AR & 0.17 (0.38) & 0.26 (0.44) & 0.29 (0.46) & 0.26 (0.44)\\
WizardLM-7B-V1.0 & SH & 0.15 (0.36) & 0.12 (0.33) & 0.11 (0.31) & 0.2 (0.4)\\
WizardLM-7B-V1.0 & SR & 0.16 (0.37) & 0.15 (0.36) & 0.16 (0.37) & 0.25 (0.43)\\

Llama-2-7b-hf & AR & 0.11 (0.32) & 0.11 (0.32) & 0.13 (0.34) & 0.16 (0.37)\\
Llama-2-7b-hf & SH & 0.13 (0.34) & 0.18 (0.38) & 0.11 (0.32) & 0.17 (0.38)\\
Llama-2-7b-hf & SR & 0.1 (0.3) & 0.07 (0.26) & 0.16 (0.37) & 0.11 (0.31)\\

\hline

guanaco-13b & AR & 0.11 (0.32) & 0.11 (0.32) & 0.11 (0.32) & 0.16 (0.37)\\
guanaco-13b & SH & 0.16 (0.37) & 0.16 (0.37) & 0.17 (0.38) & 0.16 (0.37)\\
guanaco-13b & SR & 0.14 (0.34) & 0.11 (0.32) & 0.14 (0.34) & 0.14 (0.34)\\

guanaco-33b-merged & AR & 0.13 (0.34) &  & 0.19 (0.39) & \\
guanaco-33b-merged & SH & 0.24 (0.43) &  &   & \\
guanaco-33b-merged & SR & 0.16 (0.37) &  &  & \\

mpt-30b & AR & 0.1 (0.3) & 0.17 (0.38) & 0.09 (0.28) & 0.1 (0.3)\\
mpt-30b & SH & 0.17 (0.38) & 0.16 (0.37) & 0.14 (0.34) & 0.12 (0.33)\\
mpt-30b & SR & 0.08 (0.27) & 0.13 (0.34) & 0.07 (0.26) & 0.1 (0.3)\\

WizardLM-13B-V1.2 & AR & 0.13 (0.34) & 0.24 (0.43) & 0.17 (0.38) & 0.13 (0.34)\\
WizardLM-13B-V1.2 & SH & 0.13 (0.34) & 0.16 (0.37) & 0.1 (0.3) & 0.19 (0.39)\\
WizardLM-13B-V1.2 & SR & 0.15 (0.36) & 0.18 (0.38) & 0.14 (0.35) & 0.2 (0.4)\\

WizardLM-30B-V1.0 & AR & 0.11 (0.32) & 0.17 (0.38) & 0.14 (0.35) & 0.11 (0.32)\\
WizardLM-30B-V1.0 & SR & 0.13 (0.34) & 0.14 (0.34) & 0.18 (0.38) & 0.13 (0.34)\\

falcon-40b & AR & 0.1 (0.3) & 0.21 (0.41) & 0.09 (0.28) & \\
falcon-40b & SR & 0.13 (0.34) & 0.14 (0.34) & 0.13 (0.34) & 0.14 (0.34)\\

Llama-2-13b-hf & AR & 0.07 (0.26) & 0.14 (0.35) & 0.06 (0.23) & 0.09 (0.28)\\
Llama-2-13b-hf & SH & 0.06 (0.25) & 0.06 (0.25) & 0.06 (0.23) & 0.09 (0.29)\\
Llama-2-13b-hf & SR & 0.1 (0.3) & 0.11 (0.31) & 0.12 (0.33) & 0.12 (0.33)\\

\hline
\end{tabular}
\caption{Accuracy metrics for all models tested. mean(sd)}
\label{tab:additional-models-lm}
\end{table*}

\begin{table*}
\centering
\begin{tabular}{llllll}
\hline
\textbf{Model} & \textbf{Cond} & \textbf{Classic} &\textbf{Front} & \textbf{Back} & \textbf{Both} \\
\hline

bloom-7b1 & AR & 0.13 (0.34) & 0.07 (0.26) & 0.2 (0.4) & 0.14 (0.35)\\
bloom-7b1 & SH & 0.18 (0.38) & 0.21 (0.41) & 0.16 (0.37) & 0.19 (0.39)\\
bloom-7b1 & SR & 0.29 (0.45) & 0.16 (0.37) & 0.18 (0.38) & 0.21 (0.41)\\
falcon-7b & AR & 0.13 (0.34) & 0.17 (0.38) & 0.16 (0.37) & 0.09 (0.28)\\
falcon-7b & SH & 0.19 (0.39) & 0.18 (0.38) & 0.21 (0.41) & 0.14 (0.35)\\
falcon-7b & SR & 0.29 (0.46) & 0.21 (0.41) & 0.16 (0.37) & 0.19 (0.4)\\
guanaco-7b & AR & 0.16 (0.37) & 0.14 (0.35) & 0.14 (0.35) & 0.13 (0.34)\\
guanaco-7b & SH & 0.22 (0.42) & 0.2 (0.4) & 0.16 (0.37) & 0.19 (0.4)\\
guanaco-7b & SR & 0.21 (0.41) & 0.24 (0.43) & 0.15 (0.36) & 0.14 (0.34)\\

mpt-7b-8k & AR & 0.2 (0.4) & 0.16 (0.37) & 0.2 (0.4) & 0.11 (0.32)\\
mpt-7b-8k & SH & 0.16 (0.37) & 0.21 (0.41) & 0.21 (0.41) & 0.19 (0.4)\\
mpt-7b-8k & SR & 0.26 (0.44) & 0.21 (0.41) & 0.18 (0.38) & 0.13 (0.34)\\

Llama-2-7b-hf & AR & 0.2 (0.4) & 0.13 (0.34) & 0.19 (0.39) & 0.19 (0.39)\\
Llama-2-7b-hf & SH & 0.16 (0.37) & 0.18 (0.38) & 0.16 (0.37) & 0.14 (0.34)\\
Llama-2-7b-hf & SR & 0.16 (0.37) & 0.15 (0.36) & 0.16 (0.37) & 0.14 (0.34)\\

WizardLM-7B-V1.0 & AR & 0.14 (0.35) & 0.3 (0.46) & 0.19 (0.39) & 0.19 (0.39)\\
WizardLM-7B-V1.0 & SH & 0.09 (0.29) & 0.15 (0.36) & 0.11 (0.32) & 0.19 (0.4)\\
WizardLM-7B-V1.0 & SR & 0.19 (0.39) & 0.18 (0.38) & 0.16 (0.37) & 0.21 (0.41)\\
\hline
guanaco-13b & AR & 0.23 (0.42) & 0.2 (0.4) & 0.23 (0.42) & 0.17 (0.38)\\
guanaco-13b & SH & 0.17 (0.38) & 0.17 (0.38) & 0.22 (0.42) & 0.21 (0.41)\\
guanaco-13b & SR & 0.19 (0.4) & 0.14 (0.35) & 0.14 (0.35) & 0.19 (0.4)\\
guanaco-33b-merged & SH & 0.24 (0.43) &  &  & \\
guanaco-33b-merged & SR & 0.19 (0.4) &  &  & \\

mpt-30b & AR &0.2 (0.4) & 0.19 (0.39) & 0.19 (0.39) & 0.16 (0.37)\\
mpt-30b & SH & 0.17 (0.38) & 0.16 (0.37) & 0.14 (0.34) & 0.12 (0.33)\\
mpt-30b & SR & 0.17 (0.38) & 0.16 (0.37) & 0.14 (0.34) & 0.2 (0.4)\\

falcon-40b & AR &0.17 (0.38) & 0.19 (0.39) & 0.23 (0.42) & \\
falcon-40b & SR & 0.16 (0.37) & 0.19 (0.39) & 0.1 (0.3) & 0.11 (0.32)\\

Llama-2-13b-hf & AR & 0.24 (0.43) & 0.11 (0.32) & 0.16 (0.37) & 0.16 (0.37)\\
Llama-2-13b-hf & SH & 0.24 (0.43) & 0.2 (0.4) & 0.23 (0.42) & 0.22 (0.42)\\
Llama-2-13b-hf & SR & 0.21 (0.41) & 0.14 (0.35) & 0.19 (0.4) & 0.16 (0.37)\\
WizardLM-13B-V1.2 & AR & 0.17 (0.38) & 0.27 (0.45) & 0.21 (0.41) & 0.2 (0.4)\\
WizardLM-13B-V1.2 & SH & 0.12 (0.33) & 0.15 (0.36) & 0.12 (0.33) & 0.23 (0.42)\\
WizardLM-13B-V1.2 & SR & 0.15 (0.36) & 0.14 (0.34) & 0.14 (0.34) & 0.15 (0.36)\\

WizardLM-30B-V1.0 & AR & 0.11 (0.32) & 0.17 (0.38) & 0.14 (0.35) & 0.11 (0.32)\\
WizardLM-30B-V1.0 & SH & 0.1 (0.3) & 0.19 (0.4) & 0.12 (0.33) & 0.13 (0.34)\\
WizardLM-30B-V1.0 & SR & 0.14 (0.34) & 0.19 (0.39) & 0.11 (0.32) & 0.14 (0.35)\\

\hline
\end{tabular}
\caption{DCPMI metrics for all models tested. mean(sd)}
\label{tab:additional-models-dcpmi}
\end{table*}

\end{document}